\documentclass{MITcsail}
\usepackage{times}

\usepackage[colorlinks=true, citecolor=blue, linkcolor=black]{hyperref}
\usepackage{url}
\usepackage[utf8]{inputenc} 
\usepackage{booktabs}       
\usepackage{amsfonts}       
\usepackage{nicefrac}       
\usepackage{microtype}      
\usepackage{xcolor}         
\usepackage{bm}
\usepackage{amssymb}
\usepackage{amsmath}
\usepackage{graphicx}
\usepackage{epstopdf}
\usepackage{indentfirst}
\usepackage{wrapfig}
\usepackage{subfigure}
\usepackage{array}
\usepackage{color}
\usepackage{booktabs}
\usepackage{adjustbox}
\usepackage{makecell}
\usepackage{caption}
\usepackage{algorithm}
\usepackage{algorithmic}
\usepackage[utf8]{inputenc} 
\usepackage[T1]{fontenc}    
\usepackage{url}            
\usepackage{booktabs}       
\usepackage{amsfonts}       
\usepackage{nicefrac}       
\usepackage{microtype}      
\usepackage{xcolor}         
\usepackage{bm}

\usepackage{microtype}
\usepackage{graphicx}
\usepackage{amsmath}
\usepackage{amssymb}
\usepackage{mathtools}
\usepackage{amsthm}
\usepackage{enumitem}
\usepackage{array}
\usepackage{mathptmx} 
\usepackage{wrapfig}
\usepackage{makecell}
\usepackage{caption}
\usepackage{algorithm}
\usepackage{algorithmic}

\title{SafeDiffuser: Safe Planning with Diffusion Probabilistic Models}

\author
{Wei Xiao\footnote{Correspondence E-mail: weixy@mit.edu}, Tsun-Hsuan Wang, Chuang Gan, Daniela Rus\\
\vspace{1em} 
\normalfont{\small Massachusetts Institute of Technology (MIT)}
}

\begin{document}

\maketitle
\thispagestyle{firstpagestyle} 

\begin{abstract}
Diffusion model-based approaches have shown promise in data-driven planning, but there are no safety guarantees,  thus making it hard to be applied for safety-critical applications. To address these challenges, we propose a new method, called SafeDiffuser, to ensure diffusion probabilistic models satisfy specifications by using a class of control barrier functions. The key idea of our approach is to embed the proposed finite-time diffusion invariance into the denoising diffusion procedure, which enables trustworthy diffusion data generation. Moreover, we demonstrate that our finite-time diffusion invariance method through generative models not only maintains generalization performance but also creates robustness in safe data generation. We test our method on a series of safe planning tasks, including maze path generation, legged robot locomotion, and 3D space manipulation, with results showing the advantages of robustness and guarantees over vanilla diffusion models\footnote{\color{blue} Videos can be found in the anonymous website: \url{https://safediffuser.github.io/safediffuser/}}. 
\end{abstract}

\section{Introduction}
\begin{wrapfigure}[14]{R}
{0.7\linewidth}
\vspace{-13.5mm}
	\centering
	\includegraphics[width=1\linewidth]{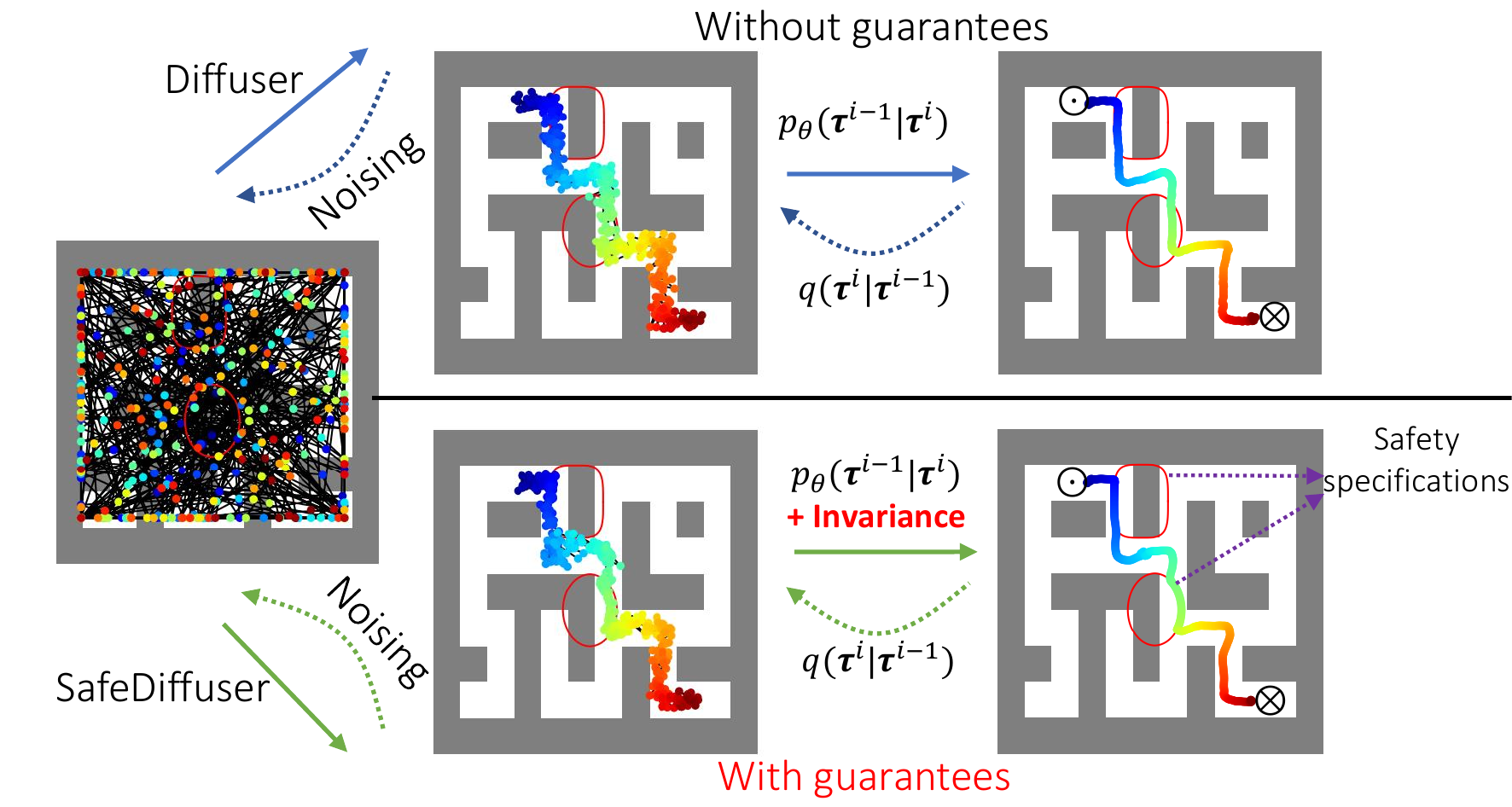}
 \vspace{-6mm}
	\caption{Our proposed SafeDiffuser (lower) generates safe trajectories with guarantees, while the diffuser (upper) fails (from $\bigodot$ to $\bigotimes$). }
	\label{fig:teaser}%
\end{wrapfigure} 

 Data-driven approaches have received increasing attentions due to their representation flexibility. Diffusion models \cite{sohl2015deep} \cite{ho2020denoising} are data-driven generative models whose primary applications are in image generations  \cite{dhariwal2021diffusion} \cite{du2020compositional} \cite{saharia2022photorealistic}.  Recently, diffusion models, termed diffusers  \cite{janner2022planning}, have shown promise in trajectory planning for a variety of robotic tasks. Diffusers enable flexible behavior synthesis that makes it well generalized in novel environments.

During inference, the diffuser, conditioned on the current state and objectives, starts from Gaussian noise to generate clean planning trajectories based on which we can get a control policy. After applying this control policy one step forward, we get a new state and run the diffusion procedure again to get a new planning trajectory. This process is repeated until the objective is achieved. However, one big challenge in this method is that there are no safety guarantees. For instance, the planning trajectory could easily violate safety constraints in the maze (as shown in Fig. \ref{fig:teaser}). This shortcoming demands a fundamental fix to diffusion models to ensure the safe generation of planning trajectories in safety-critical applications such as trustworthy policy learning and optimization.


In this paper, we propose to ensure diffusion models with specification guarantees using finite-time diffusion invariance. An invariance set is a form of specification mainly consisting of safety constraints in planning tasks. 
 We ensure that diffusion models are invariant to uncertainties in diffusion procedure. 
We achieve safety by combining receding horizon control with stable diffusion. In receding horizon control, we compute safe paths incrementally. The key insight is to replace each path computation with diffusion-based path generation, allowing a broader exploration of the path space and makes it relatively easy to include additional constraints. The computed path is combined with simulation to validate that it can be safely actuated. 

To ensure diffusers with specifications guarantees, we first find diffusion dynamics for the denoising diffusion procedure. Then, we use Lyapunov-based methods with forward invariance properties such as a class of control barrier functions (CBFs) \cite{ames2017control} \cite{Glotfelter2017} \cite{nguyen2016exponential} \cite{Xiao2019}, to formally guarantee the satisfaction of specifications at the end of the diffusion procedure. CBFs works well in planning time using robot dynamics. However, doing this in diffusion models poses extra challenges since the generated data is not directly associated with robot dynamics which makes the use of CBFs non-trivial. As oppose to existing literature, 1. we propose to embed invariance into the diffusion time for diffusers. Thus, finite-time invariance is required in diffusers since specifications are usually violated as the trajectory is initially Gaussian noise. 2. We propose to add diffusion time components in invariance to address local trap problems that are prominent in planning. 3. We propose a quadratic program approach to incorporate finite-time diffusion invariance into the diffusion to maximally preserve the performance.


In summary, we make the following \textbf{new contributions}:

\begin{itemize}
[align=right,itemindent=0em,labelsep=2pt,labelwidth=1em,leftmargin=*,itemsep=0em]
    \item We propose formal guarantees for diffusion probabilistic models via control-theoretic invariance.
    \item We propose a novel notion of finite-time diffusion invariance, and use a class of CBFs to incorporate it into the diffusion time of the procedure. We proposed three different safe diffusers, and show how we may address the local trap problem from specifications that are prominent in planning tasks.
    \item We demonstrate the effectiveness of our method on a variety of planning tasks using diffusion models, including safe planning in maze, robot locomotion and manipulation. 
\end{itemize}

\section{Preliminaries}
In this section, we provide background on diffusion models and forward invariance in control theory.

\textbf{Diffusion Probabilistic Models.} Diffusion probabilistic models \cite{sohl2015deep} \cite{ho2020denoising} \cite{janner2022planning} are latent variable models representing a data generation process as an iterative denoising procedure $p_{\theta}(\bm \tau^{i-1}|\bm \tau^i)$, $i \in\{1,\dots, N\}$, where $\bm \tau^1, \dots, \bm \tau^N$ are latent variables of the same dimensionality of the clean (noiseless) data $\bm \tau^0 \sim q(\bm \tau^0)$, and $N$ is the total denoising steps. This denoising procedure is the reverse of a forward diffusion process $q(\bm \tau^{i}|\bm \tau^{i-1})$ that gradually corrupts the clean data by adding noise.  The denoising data generation is denoted by
\begin{equation} \label{eqn:diff}
    p_{\theta}(\bm \tau^0) = \int p_{\theta}(\bm \tau^{0:N})d \bm \tau^{1:N} = \int p(\bm \tau^N) \prod_{i = 1}^{N}p_{\theta}(\bm \tau^{i-1}|\bm \tau^i)  d \bm \tau^{1:N},
\end{equation}
where $p(\bm \tau^N)$ is a standard Gaussian prior distribution, and the joint distribution $p_{\theta}(\bm \tau^{0:N})$ is defined as a Markov chain with learned Gaussian transitions starting at $p(\bm \tau^N)$. The parameter $\theta$ is optimized by minimizing the usual variational bound on the negative log-likelihood of the reverse process: $\theta^* = \arg\min_{\theta} \mathbb{E}_{\bm \tau^0}\left[- \textbf{log } p_{\theta}(\bm \tau^0)\right].$
The forward diffusion process $q(\bm \tau^{i}|\bm \tau^{i-1})$ is usually prespecified. The reverse process is often parameterized as Gaussian with time-dependent mean and variance.

\textbf{Notations.} There are two ``times'' involved in the paper: that of the diffusion process and that of the planning horizon. We use superscripts ($i$ when unspecified) to denote the diffusion time of a trajectory (state) and subscripts ($k$ when unspecified) to denote the planning time of a state on the trajectory. For instance, $\bm \tau^0$ denotes the planning trajectory at denoising diffusion time step $0$ (i.e., a noiseless trajectory), and $\bm x_k^0$ denotes the state on the trajectory at planning time step k during the denoising diffusion time step 0 (i.e., a noiseless state). We note it as $\bm x_k = \bm x_k^0$ ($\bm \tau = \bm\tau^0$ as well) when there is no ambiguity.  Further, a trajectory $\bm \tau^i$ is defined as a planning-time sequence of discretized states, i.e., $\bm \tau^i = (\bm x_0^i, \bm x_1^i, \dots, \bm x_k^i, \dots, \bm x_H^i)$, where $H\in\mathbb{N}$ is the planning horizon. During the denoising diffusion procedure, the diffusion time changes from $N$ to $0$, while the planning time varies from $0$ to $H$.

\textbf{Forward Invariance in Control Theory.}
Consider an affine control system of the form:
\begin{equation}
\dot{\bm{x}}_t=f(\bm x_t)+g(\bm x_t)\bm u_t \label{eqn:affine}%
\end{equation}
where $\bm x_t\in\mathbb{R}^{n}$, $f:\mathbb{R}^{n}\rightarrow\mathbb{R}^{n}$
and $g:\mathbb{R}^{n}\rightarrow\mathbb{R}^{n\times q}$ are {locally}
Lipschitz, and $\bm u_t\in U\subset\mathbb{R}^{q}$, where $U$ denotes a control constraint set.  $\dot{\bm x}_t$
denotes the (planning) time derivative of state $\bm x_t$.

\begin{definition}
    (\textbf{Set invariance}):
    \label{def:forwardinv} 
    A set $C\subset\mathbb{R}^{n}$ is forward invariant for system (\ref{eqn:affine}) if its solutions for some $\bm u\in U$ starting at any $\bm x_0 \in C$ satisfy $\bm x_t\in C,$ $\forall t\geq 0$.
\end{definition}

\begin{definition}
    \label{def:class_k} 
    (\textbf{Extended class $\mathcal{K}$ function} \cite{Khalil2002}): 
    A Lipschitz continuous function $\alpha: [-b,a) \rightarrow (-\infty,\infty), b>0, a>0$ belongs to extended class $\mathcal{K}$ if it is strictly increasing and $\alpha(0)=0$.
\end{definition}

Consider a safety constraint $b(\bm x_t)\geq 0$ for system (\ref{eqn:affine}), where $b:\mathbb{R}^n\rightarrow \mathbb{R}$ is continuously differentiable, we define a safe set in the form: $C := \{\bm x_t \in \mathbb{R}^n: b(\bm x_t) \geq 0\}$.

\begin{definition}
    \label{def:hocbf}
    (\textbf{Control Barrier Function (CBF)} \cite{ames2017control}):
    A function $b:\mathbb{R}^n\rightarrow\mathbb{R}$ is a CBF if there exists an extended class $\mathcal{K}$ function $\alpha$ such that
    \begin{equation}
    \label{eqn:constraint}
    \sup_{\bm u_t\in U}[L_fb(\bm x_t) + [L_gb(\bm x_t)]\bm u_t + \alpha(b(\bm x_t))] \geq 0, 
    \end{equation}
    for all $\bm x_t\in C$. $L_{f}$ and $L_{g}$ denote Lie derivatives w.r.t. $\bm x$ along $f$ and $g$, respectively.
\end{definition}

\begin{theorem} [\cite{ames2017control}]
    \label{thm:hocbf} 
    Given a CBF $b(\bm x_t)$ from Def. \ref{def:hocbf}, if $\bm x_0 \in C$, then any Lipschitz continuous controller $\bm u_t$ that satisfies the constraint in (\ref{eqn:constraint}), $\forall t\geq 0$ renders $C$ forward invariant for system (\ref{eqn:affine}).
\end{theorem}

If we need to differentiate $b(\bm x_t)$ more than once along the dynamics (\ref{eqn:affine}) until the control $\bm u_t$ explicitly shows, we use a high-order CBF \cite{nguyen2016exponential} \cite{Xiao2019} as a general form of CBF to guarantee safety for (\ref{eqn:affine}).
In this work, we map the forward invariance in control theory to finite time \textit{diffusion invariance in diffusion models}, where we incorporate CBFs into the diffusion time as opposed to their regular applications in planning time. In addition, we show how we may address the local traps during diffusion.

\section{Safe Diffuser}

In this section, we propose three different safe diffusers to ensure the safe generation of data in diffusion, i.e., to ensure the satisfaction of specifications $b(\bm x_k)\geq 0, \forall k\in\{0, \dots, H\}$. Each of the proposed safe diffusers have their own flexibility, such as avoiding local traps in planning. We consider discretized system states in the sequel. Safety in continuous planning time can be guaranteed using a lower hierarchical control framework employing other CBFs, as in \cite{ames2017control, nguyen2016exponential, Xiao2019}.

In the denoising diffusion procedure, since the learned Gaussian transitions starts at $p(\bm x^N) \sim \mathcal{N}(0, \bm I)$, it is highly likely that specifications are initially violated, i.e., $ \exists k\in\{0, \dots, H\}, b(\bm x_k^N) < 0$. For safe data generation, we wish to have $b(\bm x_k^0) \geq 0 (\textit{i.e., }b(\bm x_k) \geq 0), \forall k\in\{0, \dots, H\}$.  Since the maximum denoising diffusion step $N$ is limited, this needs to be guaranteed in a finite diffusion time step. Therefore, we propose the finite-time diffusion invariance of the diffusion procedure as follows:
\begin{definition} [Finite-time Diffusion Invariance] If there exists $i\in\{0,\dots, N\}$ such that $b(\bm x_k^j)\geq 0, \forall k\in\{0,\dots, H\}, \forall j \leq i$, then a denoising diffusion procedure $p_{\theta}(\bm\tau^{i-1}|\bm\tau^{i}), i\in\{1,\dots, N\}$ with respect to a specification $b(\bm x_k)\geq 0, \forall k\in\{0,\dots, H\}$ is finite-time diffusion invariant.
\end{definition}

The above definition can be interpreted as that if $b(\bm x_k^N) \geq 0,k\in\{0,\dots, H\}$, then we require $b(\bm x_k^i) \geq 0,\forall i\in\{0,\dots, N\}$ (similar to the forward invariance definition as in Def. \ref{def:forwardinv}); otherwise, we require that  $b(\bm x_k^j) \geq 0,\forall j\in\{0,\dots, i\}, i\in\{0,\dots, N\}$, where $i$ is a finite diffusion time.

In the following, we propose three different methods to achieve finite-time diffusion invariance. The first method is a general form of the safe- diffuser, and the other two are variants to address local traps in planning.

\subsection{Robust-Safe Diffuser}
\begin{wrapfigure}[16]{R}
{0.5\linewidth}
\vspace{-2mm}
	\centering
	\includegraphics[width=.95\linewidth]{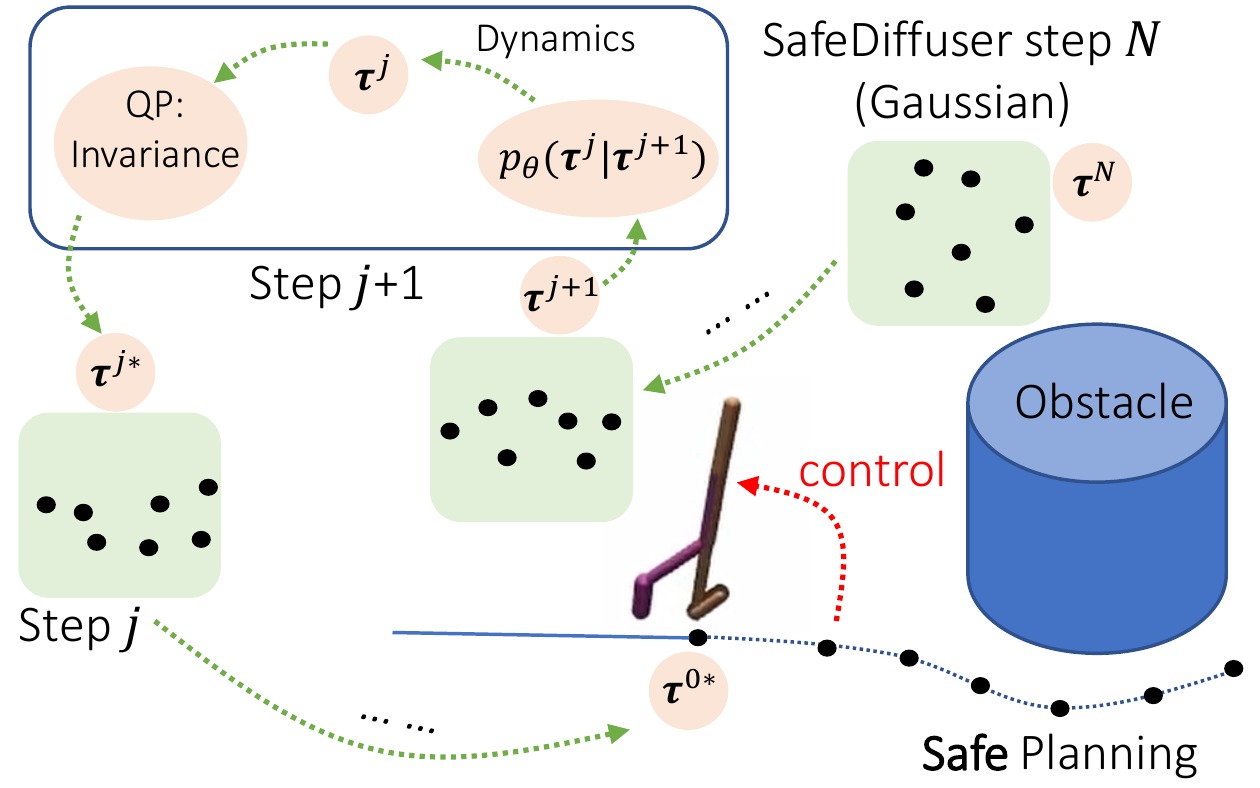}
    \caption{The proposed SafeDiffuser workflow.  SafeDiffuser performs an additional step of invariance QP solver in the diffusion dynamics to ensure safety. The final control signal is inferred from safe planning trajectories.}
	\label{fig:frame}%
\end{wrapfigure} 
The safe denoising diffusion procedure is considered at every diffusion step. Following (\ref{eqn:diff}),  the data generation at the diffusion time $j\in\{0,\dots, N-1\}$ is given by:
\begin{equation} \label{eqn:diffj}
    p_{\theta}(\bm \tau^j) = \int p(\bm \tau^N)\prod_{i = j+1}^Np_{\theta}(\bm \tau^{i-1}|\bm \tau^i) d\bm \tau^{j+1:N}
\end{equation}

A sample $\bm \tau^j, j\in\{0,\dots, N-1\}$ follows the data distribution in (\ref{eqn:diffj}), i.e., we have
\begin{equation} \label{eqn:diffj1}
    \bm \tau^j \sim p_{\theta}(\bm \tau^j).
\end{equation}

The denoising diffusion dynamics are then given by:
\begin{equation} \label{eqn:diff_dy}
    \dot {\bm \tau}^j = \lim_{\Delta \tau\rightarrow 0}\frac{\bm \tau^j - \bm \tau^{j+1}}{\Delta\tau}
\end{equation}
where $\dot{\bm \tau}$ is the (diffusion) time derivative of $\bm \tau$. $\Delta\tau > 0$ is a small enough diffusion time step length during implementations, and $\bm \tau^{j+1}$ is available from the last diffusion step. 

In order to impose finite-time diffusion invariance on the diffusion procedure,  we wish to make diffusion dynamics (\ref{eqn:diff_dy}) controllable. We reformulate (\ref{eqn:diff_dy}) as
\begin{equation} \label{eqn:diff_ctrl}
    \dot {\bm \tau}^j = \bm u^j,
\end{equation}
where $\bm u^j$ is a control variable of the same dimensionality as ${\bm \tau}^j$. 
On the other hand, we wish $\bm u^j$ to stay close to $\frac{\bm \tau^j - \bm \tau^{j+1}}{\Delta\tau}$ in order to maximally preserve the performance of the diffusion model. The above model can be rewritten in terms of each state on the trajectory $\bm \tau^j$: $\dot{\bm x}_k^j= \bm u_k^j,$
where $\bm u_k^j$ is the $k^{th}$ component of $\bm u^j$. Then, we can define CBFs to ensure the satisfaction of $b(\bm x_k^j)\geq 0$ (in finite diffusion time):
$
    h(\bm u_k^j|\bm x_k^j):= \frac{db(\bm x_k^j)}{d\bm x_k^j } \bm u_k^j + \alpha(b(\bm x_k^j)) \geq 0, k\in\{0,\dots, H\}, j\in\{0,\dots, N-1\},
$
where $\alpha(\cdot)$ is an extended class $\mathcal{K}$ function. We have the following theorem to show the finite-time diffusion invariance (proof is given in appendix):
\begin{theorem} \label{thm:diff1}
    Let the diffusion dynamics defined as in (\ref{eqn:diff_dy}) whose controllable form is defined as in (\ref{eqn:diff_ctrl}). If there exists an extended class $\mathcal{K}$ function $\alpha$ such that
    \begin{equation} \label{eqn:diff_cbf}
        h(\bm u_k^j|\bm x_k^j) \geq 0, \forall k\in\{0,\dots, H\}, \forall j\in\{0,\dots, N-1\},
    \end{equation}
    where $h(\bm u_k^j|\bm x_k^j) = \frac{db(\bm x_k^j)}{d\bm x_k^j } \bm u_k^j + \alpha(b(\bm x_k^j))$, then the diffusion procedure $p_{\theta}(\bm\tau^{i-1}|\bm\tau^{i}), i\in\{1,\dots, N\}$ is finite-time diffusion invariant with almost probability 1.
\end{theorem}

One possible issue in the robust-safe diffusion procedure is that if $b(\bm x_k^j) \geq 0$ when $j$ is close to the initial diffusion step $N$, then the state $\bm x_k^j$ can never violate the specification after diffusion step $j < N$. When there are local traps from specifications, the state $\bm x_k^j$ may get stuck there during the denoising diffusion process, which may adversely affect the performance. In order to address this issue, we propose relaxed-safe diffuser and time-varying-safe diffuser in the following subsections.

\subsection{Relaxed-Safe Diffuser}

In order to address the local trap problems imposed by specifications during the denoising diffusion procedure, we propose a variation of the robust-safe diffuser. We define the diffusion dynamics and their controllable form as in (\ref{eqn:diff_dy}) - (\ref{eqn:diff_ctrl}).  The modified versions for CBFs are in the form:
\begin{equation} \label{eqn:diff_cbf_r}
    h(\bm u_k^j, r_k^j|\bm x_k^j):= \frac{db(\bm x_k^j)}{d\bm x_k^j } \bm u_k^j + \alpha(b(\bm x_k^j)) -w_k(j)r_k^j\geq 0, k\in\{0,\dots, H\}, j\in\{0,\dots, N-1\},
\end{equation}
where $r_k^j\in\mathbb{R}$ is a relaxation variable that is to be determined (shown in the next section). $w_k(j) \geq 0$ is a diffusion time-varying weight on the relaxation variable such that it gradually decrease to 0 as $j\rightarrow 0$.

When $w_k(j)$ decreases to 0, the condition (\ref{eqn:diff_cbf_r}) becomes a hard constraint. One problem in such cases is that the diffusion performance may be adversely affected by such a hard constraint. In order to address this issue, we may run additional $N_a\in\mathbb{N}$ diffusion steps, while setting the diffusion time to 0 when $j < 0$ in the reverse process.  The corresponding CBF conditions to (\ref{eqn:diff_cbf_r}) are to change the domain of $j$ to $j\in\{-N_a,\dots, N-1\}$.
We also have the following theorem to show the finite-time diffusion invariance (proof is given in appendix):
\begin{theorem}
    Let the diffusion dynamics defined as in (\ref{eqn:diff_dy}) whose controllable form is defined as in (\ref{eqn:diff_ctrl}). If there exist an extended class $\mathcal{K}$ function $\alpha$, a large enough extra diffusion step $N_a\in\mathbb{N}$, and a time-varying weight $w_k(j)$ where $w_k(j) = 0$ for all $j \leq 0$ such that
    \begin{equation} \label{eqn:diff_rcbf}
        h(\bm u_k^j, r_k^j|\bm x_k^j) \geq 0, \forall k\in\{0,\dots, H\}, \forall j\in\{-N_a,\dots, N-1\},
    \end{equation}
    where $h(\bm u_k^j,r_k^j|\bm x_k^j) = \frac{db(\bm x_k^j)}{d\bm x_k^j } \bm u_k^j + \alpha(b(\bm x_k^j)) -w_k(j)r_k^j$, then the diffusion procedure $p_{\theta}(\bm\tau^{i-1}|\bm\tau^{i}), i\in\{-N_a,\dots, N\}$ is finite-time diffusion invariant with almost probability 1.
\end{theorem}


After the denoising diffusion procedure is done at step $-N_a$, we would set the $\bm\tau^{-N_a}$ as the output data of the diffusion model, i.e., $\bm \tau^0 = \bm\tau^{-N_a}$.

\subsection{Time-Varying-Safe Diffuser}

As an alternative to the relaxed-safe diffuser, we propose another safe diffuser called time-varying-safe diffuser in this subsection. The proposed time-varying-safe diffuser can also address the local trap issues induced by specifications.

In this case, we directly modify the specification $b(\bm x_k^j)\geq 0$ by a diffusion time-varying function $\gamma_k:j \rightarrow \mathbb{R}$ in the form:
\begin{equation} \label{eqn:spe_t}
    b(\bm x_k^j) - \gamma_k(j) \geq 0,  k\in\{0,\dots, H\}, j\in\{0, \dots, N\},
\end{equation}
where $\gamma_k(j)$ is continuously differentiable, and is defined such that $\gamma_k(N) \leq b(\bm x_k^N)$ and $\gamma_k(0) = 0$. 

The modified time-varying specification can then be enforced using CBFs: $h(\bm u_k^j|\bm x_k^j, \gamma_k(j)):= \frac{db(\bm x_k^j)}{d\bm x_k^j } \bm u_k^j -\dot\gamma_k(j)+ \alpha(b(\bm x_k^j) - \gamma_k(j)) \geq 0, k\in\{0,\dots, H\}, j\in\{0,\dots, N-1\},$
where $\dot\gamma_k(j)$ is the diffusion time derivative of $\gamma_k(j)$.
Finally, we have the following theorem to show the finite-time diffusion invariance (proof is given in appendix):
\begin{theorem} \label{thm3}
    Let the diffusion dynamics defined as in (\ref{eqn:diff_dy}) whose controllable form is defined as in (\ref{eqn:diff_ctrl}). If there exist an extended class $\mathcal{K}$ function $\alpha$ and a time-varying function $\gamma_k(j)$ where $\gamma_k(N) \leq b(\bm x_k^N)$ and $\gamma_k(0) = 0$ such that
    \begin{equation} \label{eqn:diff_cbft}
        h(\bm u_k^j|\bm x_k^j, \gamma_k(j)) \geq 0, \forall k\in\{0,\dots, H\}, \forall j\in\{0,\dots, N-1\},
    \end{equation}
    where $h(\bm u_k^j|\bm x_k^j, \gamma_k(j)) = \frac{db(\bm x_k^j)}{d\bm x_k^j } \bm u_k^j -\dot\gamma_k(j)+ \alpha(b(\bm x_k^j) - \gamma_k(j))$, then the diffusion procedure $p_{\theta}(\bm\tau^{i-1}|\bm\tau^{i}), i\in\{0,\dots, N\}$ is finite-time diffusion invariant.
\end{theorem}





\section{Enforcing Invariance in Diffuser}
\label{sec:alg}

In this section, we show how we may incorporate the three proposed invariance methods from the last section into diffusion models. Enforcing the finite-time invariance in diffusion models is equivalent to ensure the satisfaction of the conditions in Thms. \ref{thm:diff1}-\ref{thm3} in the diffusion procedure. In this section, we propose a minimum-deviation quadratic program (QP) approach to achieve that. We wish to enforce these conditions at every step of the diffusion (as shown in Fig. \ref{fig:frame}) as those states that are far from the specification boundaries $b(\bm x_k^j)=0$ can also be optimized accordingly, and thus, the model may generate more coherent trajectories.

\textbf{Enforcing Invariance for Robust-Safe (RoS) and Time-Varying-Safe Diffusers.}
During implementation, the diffusion time step length $\Delta \bm \tau$ in (\ref{eqn:diff_dy}) is chosen to be small enough, and we wish the control $\bm u^j$ to stay close to the right-hand side of (\ref{eqn:diff_dy}). Thus, we can formulate the following QP-based optimization to find the optimal control for $\bm u^j$ that satisfies the condition in Thms. \ref{thm:diff1} or \ref{thm3}:
\begin{equation} \label{eqn:qp1}
\begin{aligned}
    \bm u^{j*} &= \arg\min_{\bm u^j} ||\bm u^j - \frac{\bm \tau^j - \bm \tau^{j+1}}{\Delta\tau}||^2, \text{ s.t., } (\ref{eqn:diff_cbf}) \text{ if RoS diffuser }  \text{ else s.t., } (\ref{eqn:diff_cbft}), 
\end{aligned}
\end{equation}
where $||\cdot||$ denotes the 2-norm of a vector. If we have more than one specification, we can add the corresponding conditions in Thm. \ref{thm:diff1} for each of them to the above QP. After we solve the above QP and get $\bm u^{j*}$, we update (\ref{eqn:diff_ctrl}) by setting $\bm u^j = \bm u^{j*}$ within the time step and get a new state for the diffusion procedure. Note that all of these happen at the end of each diffusion step.

\begin{algorithm}[tb]
   \caption{Enforcing invariance in diffusion models}
   \label{alg:pp}
\begin{algorithmic}
   \STATE {\bfseries Input:} the last trajectory of diffusion $\bm \tau^{j+1}$ at diffusion step $j\in\{0,\dots,N\}$
   \STATE {\bfseries Output:} safe diffusion state $\bm \tau^{j*}$.
   \STATE (a) Run diffusion procedure (\ref{eqn:diffj}) and sample (\ref{eqn:diffj1}) as usual at step $j$ and get $\bm \tau^j$.
   \STATE (b) Find diffusion dynamics as in (\ref{eqn:diff_dy}) - (\ref{eqn:diff_ctrl}).
   \IF{\textit{Robust-safe diffuser}}
   \STATE Formulate the QP (\ref{eqn:qp1}), solve it and get $\bm u^{j*}$.
   \ELSIF{Relaxed-safe diffuser}
   \STATE Define the time-varying weight $w_k(j)$ in (\ref{eqn:diff_cbf_r}), formulate the QP (\ref{eqn:qp2}), solve it and get $\bm u^{j*}, r^{j*}$.
    \ELSE
    \STATE Design the time-varying function $\gamma_k(j)$ in (\ref{eqn:spe_t}), formulate the QP (\ref{eqn:qp1}), solve it and get $\bm u^{j*}$.
   \ENDIF
   \STATE (c) Update dynamics (\ref{eqn:diff_ctrl}) with $\bm u^j = \bm u^{j*}$ and get $\bm \tau^{j*}$. Finally, $\bm \tau^j\leftarrow\bm \tau^{j*}$.
\end{algorithmic}
\end{algorithm}

\textbf{Enforcing Invariance for Relaxed-Safe Diffuser.}
In this case, since we have relaxation variables for each of the safety specification, we wish to minimize these relaxations in the cost function to drive all the state towards the satisfaction of specifications. In other words, we have the following QP:
\begin{equation} \label{eqn:qp2}
\begin{aligned}
    \bm u^{j*}, r^{j*} &= \arg\min_{\bm u^j, r^j} ||\bm u^j - \frac{\bm \tau^j - \bm \tau^{j+1}}{\Delta\tau}||^2 + ||r^{j}||^2, \text{ s.t., } (\ref{eqn:diff_rcbf}),
\end{aligned}
\end{equation}
where $r^j$ is the concatenation of $r_k^j$ for all $k\in\{0,\dots, H\}$. As an alternative, all the constraints above may share the same relaxation variable, i.e., the dimension of $r^j$ is only one. After we solve the above QP and get $\bm u^{j*}$, we update (\ref{eqn:diff_ctrl}) by setting $\bm u^j = \bm u^{j*}$ within the time step and get a new state. 



\textbf{Complexity of enforcing invariance} The computational complexity of a QP is $\mathcal{O}(q^3)$, where $q$ is the dimension of the decision variable.  When there is a set $S$ of specifications, we just add the corresponding constraints for each specification the QP. The complexity of the three proposed safe diffuser are similar. 

The algorithm for enforcing invariance is straight forward, which includes the construction of proper conditions, the solving of QP, and the update of diffusion state. We summary the algorithm in Alg. \ref{alg:pp}.

\begin{figure}[t]
    \centering
    \subfigure{\includegraphics[width=0.24\linewidth]{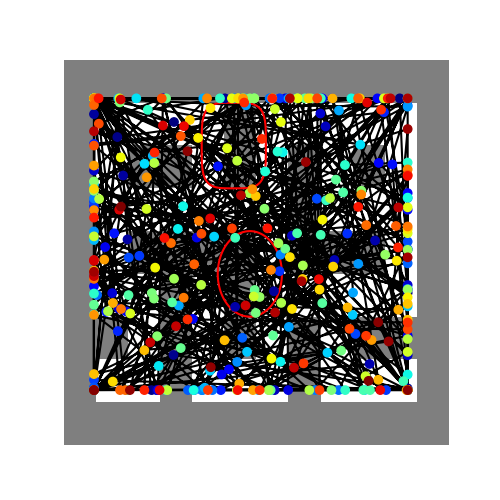}}
    \subfigure{\includegraphics[width=0.24\linewidth]{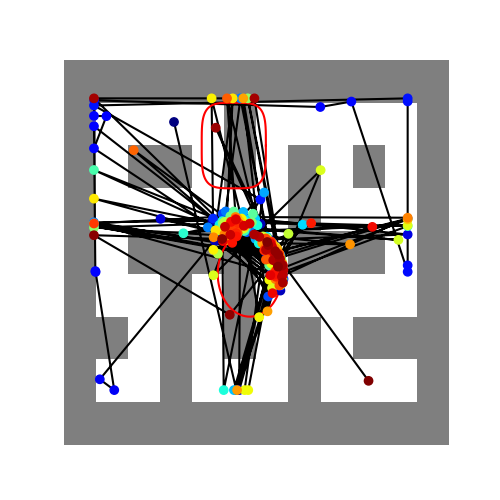}}
    \subfigure{\includegraphics[width=0.24\linewidth]{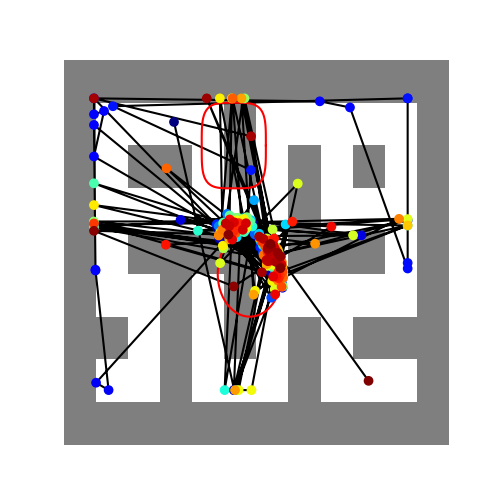}}
    \subfigure{\includegraphics[width=0.24\linewidth]{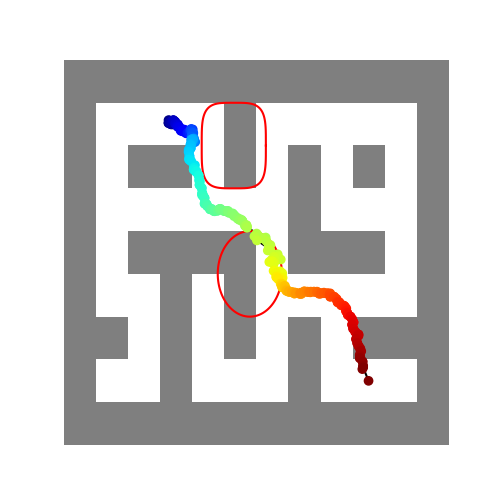}}
    \vspace{-3mm}
    \caption{Maze planning (blue to red) denoising diffusion procedure with classifier-based guidance (Left to right: diffusion time steps 256, 4, 3, -50, respectively). Red ellipse and super-ellipse (outside) denote safe specifications. The safe classifier-based guidance approach adversely affects the diffusion procedure without guarantees.
    }
    \label{fig:maze_gd}
    \vspace{-5mm}
\end{figure}

\section{Experiments}
\label{sec:exp}

\begin{table}[ht]
\caption{Maze safe planning comparisons with benchmarks.  Items are short for satisfaction of simple specifications (S-SPEC) and complex specifications (C-SPEC), score of planning tasks (SCORE), computation time at each diffusion step (TIME) in seconds, respectively. In the method column, items are short for robust-safe diffuser (RoS-DIFFUSER), relaxed-safe diffuser (ReS-DIFFUSER), and time-varying-safe diffuser (TVS-DIFFUSER), respectively. The classifier guidance$-\epsilon$ method applies (safe) gradient to the model when the state is $\epsilon>0$ close to the boundary.}
\label{tab:comp_mz}
\vskip 0.05in
\begin{center}
\begin{small}
\resizebox{0.8\textwidth}{!}{
\begin{sc}
\begin{tabular}{p{4.8cm}<{\centering}p{1.3cm}<{\centering}<{\centering}p{1.3cm}<{\centering}p{1.8cm}<{\centering}p{0.8cm}<{\centering}r}
\toprule
Method & S-spec($\uparrow\& \geq 0$)  & C-spec($\uparrow\& \geq 0$)     & Score ($\uparrow$)  & time  \\
\midrule
Diffuser (baseline)   \cite{janner2022planning}
& -0.983 & -0.894   &$1.016 {\small \pm 0.712}$  & 0.007 \\
\midrule
Truncate \cite{gym} 
     & $-1.192e^{-7}$ & -0.759  &$0.754 {\small \pm 0.779}$  &  0.024   \\
     
Classifier guidance \cite{dhariwal2021diffusion} 
& -0.789 & -0.979  &$0.502 {\small \pm 0.328}$   &  0.053   \\

Classifier guidance$-\epsilon$ \cite{dhariwal2021diffusion}
& -0.853 & -0.995  &$0.470 {\small \pm 0.366}$   &  0.061    \\

RoS-diffuser (Ours) 
& $-2.384e^{-7}$ & $-5.960e^{-7}$   & $0.770 {\small \pm 0.782}$  &0.106   \\

ReS-diffuser (Ours)
  & $-2.384e^{-7}$ & $-4.768e^{-7}$  &$0.762 {\small \pm 0.746}$  &  0.107    \\
TVS-diffuser (Ours)
  & $-2.384e^{-7}$ & $-5.364e^{-7}$  &$0.806 {\small \pm 0.783}$   &  0.107  
\\\bottomrule
\end{tabular}
\end{sc}
}
\end{small}
\end{center}
\vskip -0.1in
\end{table}

We set up experiments to answer the following questions:
\vspace{-2ex}
\begin{itemize}
[align=right,itemindent=0em,labelsep=2pt,labelwidth=1em,leftmargin=*,itemsep=0em] 

    \item Does our method match the theoretical potential in various tasks quantitatively and qualitatively?
    \item How does our method compare with state-of-the-art approaches in enforcing safety specifications?
    \item How does our proposed method affect the performance of diffusion  under guaranteed specifications? 
\end{itemize}

\subsection{Safe Planning in Maze}
In this experiment, we aim to impose trajectory constraints on the planning path of a maze. The training data  is publicly available from \cite{janner2022planning}, in which initial positions and destinations in maze are randomly generated. The diffusion model is conditioned on the initial positions and destinations.


\begin{figure}[t]
    \centering
    \subfigure{\includegraphics[width=0.24\linewidth]{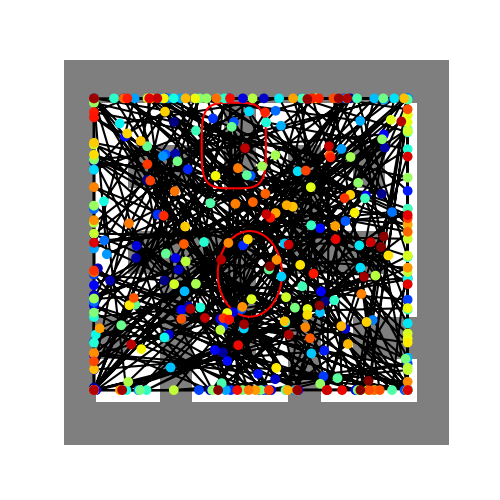}}
    \subfigure{\includegraphics[width=0.24\linewidth]{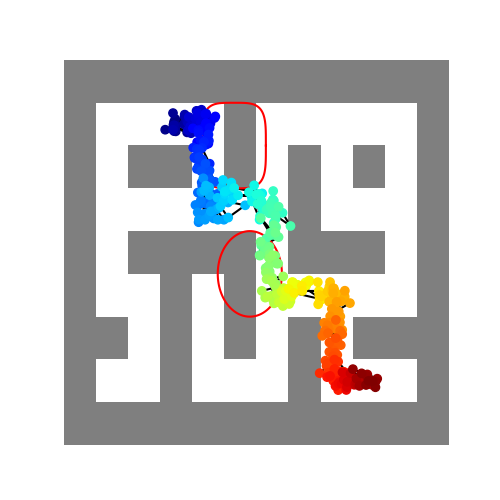}}
    \subfigure{\includegraphics[width=0.24\linewidth]{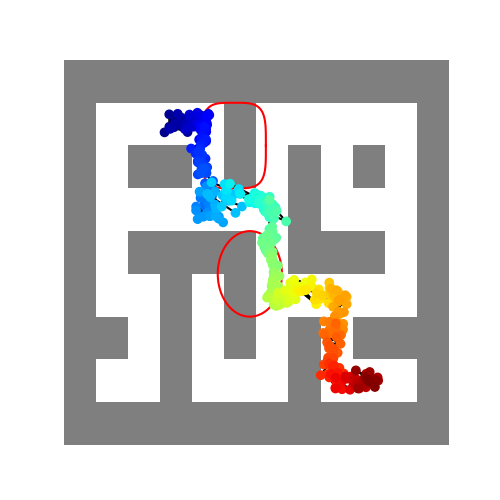}}
    \subfigure{\includegraphics[width=0.24\linewidth]{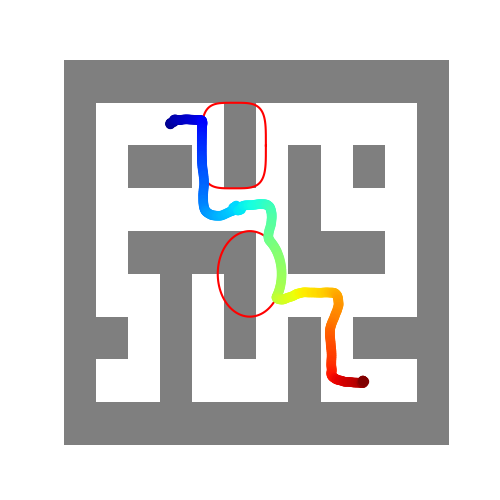}}
    \vspace{-3mm}
    \caption{Maze planning (blue to red) denoising diffusion procedure with the proposed time-varying safe diffuser (Left to right: diffusion time steps 256, 4, 3, -50, respectively). Red ellipse and super-ellipse (outside) denote safe specifications.  The proposed time-varying safe diffuser can guarantee specifications at the end of diffusion while not significantly affecting the diffusion procedure.
    }
    \label{fig:maze_cbf}
    \vspace{-5mm}
\end{figure}

The diffuser cannot guarantee the satisfaction of any specifications, as shown in Fig. \ref{fig:teaser}. When using classifier-based guidance in diffusion for safety specifications, the performance could be significantly affected (Fig. \ref{fig:maze_gd}). As a result, the generated trajectory will largely deviates from the desired one with no safety. The proposed robust-safe diffuser (RoS-diffuser), relaxed-safe diffuser (ReS-diffuser), and time-varying-safe diffuser (TVS-diffuser) can all guarantee the satisfaction of specifications, even when the specifications are complex (as long as they are differentiable), as shown in Table \ref{tab:comp_mz}. The satisfaction scores of the proposed methods are not strictly positive, and this may be due to the computation errors or inter-sampling effect as even the truncation method cannot strictly satisfy the simple specifications. The proposed methods can also maximally preserve the performance of diffusion models, as shown by the score in Table \ref{tab:comp_mz}, as well as in Fig. \ref{fig:maze_cbf}. The ReS-diffuser and TVS-diffuser can address the local trap problem from specifications, as shown by figures in appendix.

\subsection{Safe Planning for Robot Locomotion}

For robot locomotion (in MuJoCo), we wish the robot to avoid collisions with obstacles, such as the roof. In this case, since there is no local trap problem, we only consider robust-safe diffuser (RoS-diffuser). Others work similarly. The training data set is publicly available from \cite{janner2022planning}.

\begin{table}[ht]
\caption{Robot safe planning comparisons with benchmarks.  Items are short for satisfaction of simple specifications (S-SPEC), satisfaction of complex specifications (C-SPEC), score of planning tasks (SCORE), computation time at each diffusion step (TIME) in seconds, respectively. 
}
\label{tab:robot}
\vskip 0.05in
\begin{center}
\begin{small}
\resizebox{0.95\textwidth}{!}{
\begin{sc}
\begin{tabular}{p{1.4cm}<{\centering}p{4.8cm}<{\centering}p{1cm}<{\centering}<{\centering}p{1.3cm}<{\centering}p{1.8cm}<{\centering}p{0.8cm}<{\centering}r}
\toprule
Experiment & Method & S-spec($\uparrow\& \geq 0$)  & C-spec($\uparrow\& \geq 0$)     & Score ($\uparrow$)   & time  \\
\midrule
& Diffuser (baseline)   \cite{janner2022planning}
& -9.375 & -4.891   &$0.346 {\small \pm 0.106}$  & 0.037 \\

Walker2D&Truncate \cite{gym} 
     & $0.0$ & $\times$  &$0.286 {\small \pm 0.180}$  &  0.105   \\
     
&Classifier guidance \cite{dhariwal2021diffusion}
& -0.575 & -0.326  &$0.208 {\small \pm 0.140}$   &  0.053   \\

&RoS-diffuser (Ours) 
& $0.0$ & $-6.706e^{-8}$   & $0.312 {\small \pm 0.782}$  &0.183   \\

\midrule

& Diffuser (baseline)   \cite{janner2022planning}
& -2.180 & -1.862   &$0.455 {\small \pm 0.038}$  & 0.038 \\

Hopper&Truncate \cite{gym}  
     & $0.0$ & $\times$  &$0.436 {\small \pm 0.067}$  &  0.046   \\
     
&Classifier guidance \cite{dhariwal2021diffusion} 
& -0.894 & -0.524  &$0.478 {\small \pm 0.038}$   &  0.047   \\

&RoS-diffuser (Ours) 
& $0.0$ & $-5.960e^{-8}$   & $0.430 {\small \pm 0.040}$  &0.170   
\\\bottomrule
\end{tabular}
\end{sc}
}
\end{small}
\end{center}
\vskip -0.1in
\end{table}

\begin{figure}[t]
    \centering
    \subfigure{\includegraphics[width=0.995\linewidth]{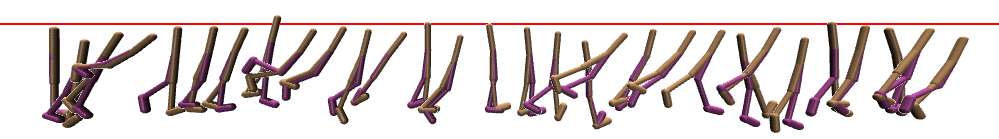} {\color{blue}$\downarrow$}}\vspace{-7mm}
    \subfigure{\includegraphics[width=1\linewidth]{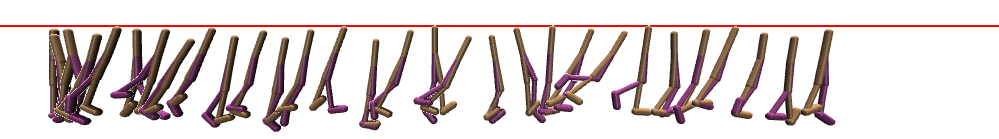}{\color{blue}$ \downarrow$}}\vspace{-7mm}
    \subfigure{\includegraphics[width=1\linewidth]{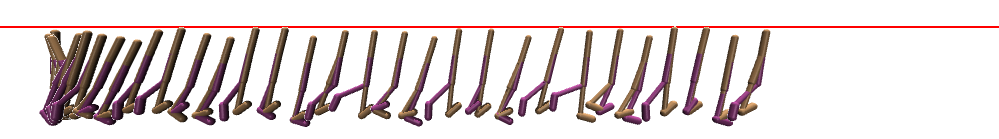}}
    \vspace{-5mm}
    \caption{Walker2D planning denoising diffusion procedure with the proposed robust-safe diffuser (Up to down: diffusion time steps 20, 10, 0, respectively). The red line denotes the roof the walker needs to safely avoid during locomotion (safety specifications). Safety is violated at step 20 since the trajectory is initially Gaussian noise, but is eventually guaranteed (step 0). 
    }
    \label{fig:walker_cbf}
    \vspace{-3mm}
\end{figure}

As expected, collisions with roof are very likely to happen in walker and hopper using the diffuser since there are no guarantees, as shown in Table \ref{tab:robot}. The truncation method can work for simple specifications (S-spec), but not for complex specifications (C-spec). The classifier-based guidance can improve the satisfaction of specifications, but with no guarantees. Collision-free is guaranteed using the proposed Ros-diffuser, and one example of diffusion procedure is shown in Fig. \ref{fig:walker_cbf}. 

\subsection{Safe Planning for Manipulation}

For manipulation (in Pybullet), the diffusion models generate joint trajectories (as controls) for the robot, which are conditioned on the locations of the objects to grasp and place. The training data set is publicly available from \cite{janner2022planning}. Specifications are joint limitations to avoid collision in joint space.

\begin{table}[ht]
\caption{Manipulation safe planning comparisons with benchmarks.  Items are short for satisfaction of simple specifications (S-SPEC), satisfaction of complex specifications (C-SPEC), reward of planning tasks (REWARD), computation time at each diffusion step (TIME) in seconds, respectively. In the method column, the item is short for robust-safe diffuser (RoS-DIFFUSER).}
\label{tab:mani}
\vskip 0.05in
\begin{center}
\begin{small}
\resizebox{0.95\textwidth}{!}{
\begin{sc}
\begin{tabular}{p{4.8cm}<{\centering}p{1.3cm}<{\centering}<{\centering}p{1.3cm}<{\centering}p{1.8cm}<{\centering}p{0.8cm}<{\centering}r}
\toprule
Method & S-spec($\uparrow\& \geq 0$)  & C-spec($\uparrow\& \geq 0$)     & Reward ($\uparrow$)   & time  \\
\midrule
Diffuser (baseline)  \cite{janner2022planning} 
& -0.057 & -0.065   &$0.650 {\small \pm 0.107}$  & 0.038 \\
\midrule
Truncate \cite{gym} 
     & $1.631e^{-8}$ & $\times$  &$0.575 {\small \pm 0.112}$  &  0.069   \\
     
Classifier guidance \cite{dhariwal2021diffusion} 
& -0.050 & -0.053  &$0.800 {\small \pm 0.328}$   &  0.075   \\

RoS-diffuser (Ours) 
& $0.072$ & $0.069$   & $0.925 {\small \pm 0.107}$  &0.088   
\\\bottomrule
\end{tabular}
\end{sc}
}
\end{small}
\end{center}
\vskip -0.1in
\end{table}

\begin{figure}[t]
    \centering
    \subfigure{\includegraphics[width=0.24\linewidth]{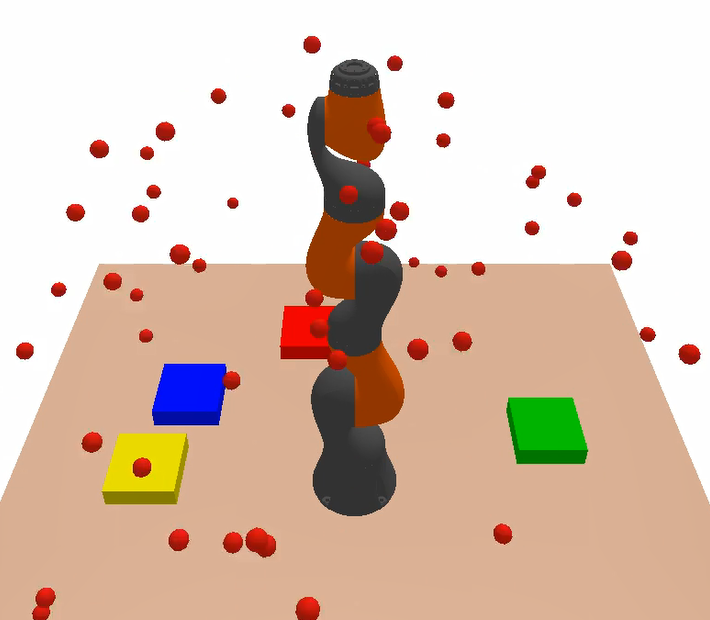}}
     \subfigure{\includegraphics[width=0.24\linewidth]{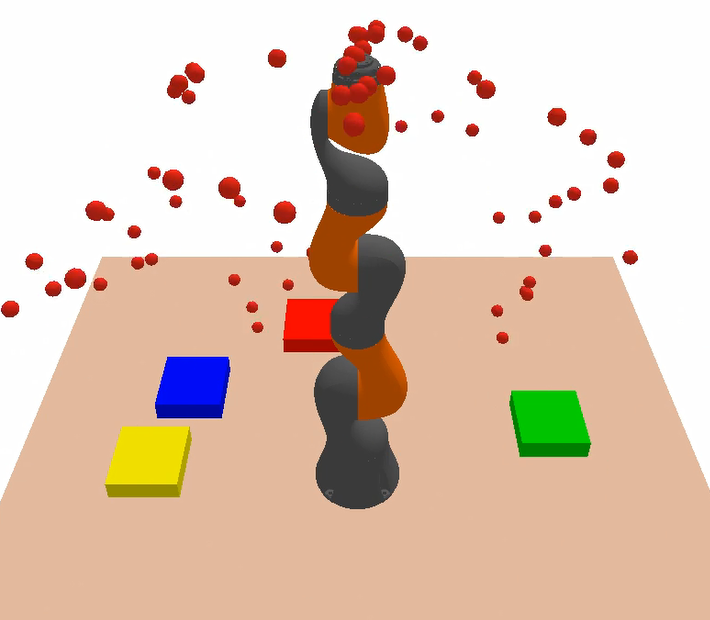}}
    \subfigure{\includegraphics[width=0.24\linewidth]{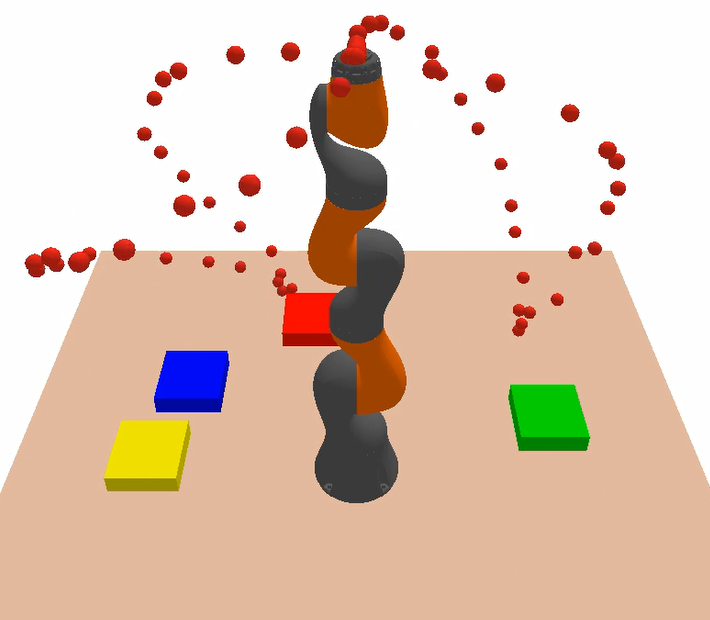}}
    \subfigure{\includegraphics[width=0.24\linewidth]{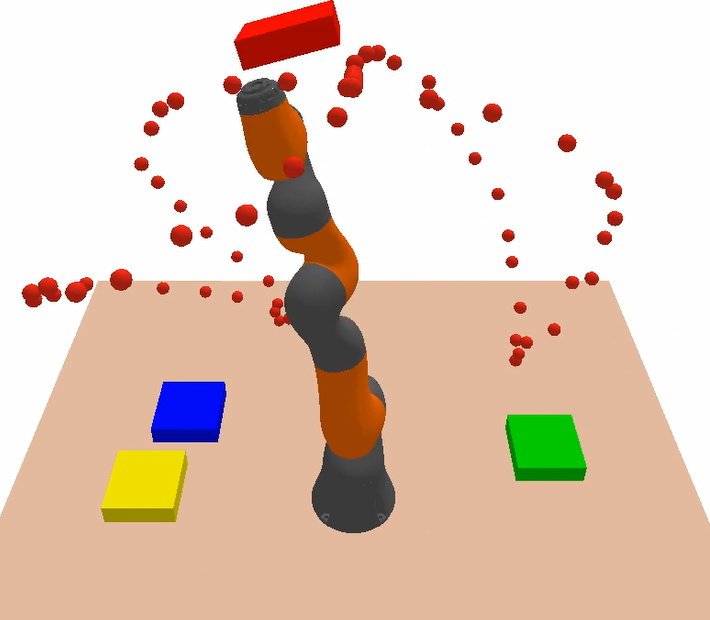}}
    \vspace{-3mm}
    \caption{Manipulation planning denoising diffusion procedure with the proposed robust-safe diffuser (Left to right: diffusion time steps 1000, 100, 0, and execution time step 100, respectively). The red dots denote the planning trajectory of the end-effector. 
    }
    \label{fig:kuka_cbf}
    \vspace{-5mm}
\end{figure}

In this case, the truncation method still fails to work for complex specifications (speed-dependent joint limitations). 
Our proposed RoS-diffuser can work for all specifications as long as they are differentiable. An interesting observation is that the proposed RoS-diffuser can even improve the performance (reward) of  diffusion models in this case, as shown in Table \ref{tab:mani}. This may be due to the fact that the satisfaction of joint limitations can avoid collision in the joint space of the robot as Pybullet is a physical simulator. The computation time of the proposed RoS-diffuser is comparable to other methods. An illustration for the safe diffusion and manipulation procedure is shown in Fig. \ref{fig:kuka_cbf}.

\vspace{-1.5ex}
\section{Related Works}
\vspace{-1.5ex}
\noindent \textbf{Diffusion models and planning} Diffusion models \cite{sohl2015deep} \cite{ho2020denoising} are data-driven generative modeling tools, widely used in applications to image generations \cite{dhariwal2021diffusion} \cite{du2020compositional}, in planning \cite{hafner2019learning} \cite{janner2021offline} \cite{ozair2021vector} \cite{janner2022planning}, and in language \cite{saharia2022photorealistic} \cite{liu2023audioldm}. Generative models are combined with reinforcement learning to explore dynamic models in the form of convolutional U-networks \cite{kaiser2019model}, stochastic recurrent networks \cite{ke2019modeling}, neural ODEs \cite{du2020model},  generative
adversarial networks \cite{eysenbach2022mismatched},  neural radiance fields \cite{li20223d}, and transformers \cite{chen2022transdreamer}. Further, planning tasks are becoming increasingly important for diffusion models \cite{lambert2021learning} \cite{ozair2021vector} \cite{janner2022planning} as they can generalize well in all kinds of robotic problems.
However, there are no methods to equip diffusion models with specification guarantees, which is especially important for safety-critical applications. Here, we address this issue using the proposed finite-time diffusion invariance. 

\noindent \textbf{Set invariance and CBFs.} An invariant set has been widely used to represent the safe behavior of dynamical systems \cite{preindl2016robust} \cite{rakovic2005invariant} \cite{ames2017control} \cite{Glotfelter2017} \cite{Xiao2021bnet}. In the state of the art of control, Control Barrier Functions (CBFs) are also widely used to prove set
invariance \cite{Aubin2009}, \cite{Prajna2007}, \cite{Wisniewski2013}. CBFs can be traced back to optimization problems
\cite{Boyd2004}, and are Lyapunov-like functions
\cite{Wieland2007}. For time-varying systems, CBFs can also be adapted accordingly \cite{Lindemann2018}.  Existing CBF approaches are usually applied in planning time since they are closely coupled with system dynamics. There are few studies of CBFs in other space, such as the diffusion time horizon in diffusion models. Our work addresses all these limitations.

\noindent \textbf{Guarantees in neural networks.}
Differentiable optimization methods show promise for neural network controllers with guarantees \cite{pereira2020, Amos2018, Xiao2021bnet}. They are usually served as a layer (filter) in the neural networks. In \cite{Amos2017}, a differentiable quadratic program (QP) layer, called OptNet, was introduced. OptNet with CBFs has been used in neural networks as a filter for safe controls \cite{pereira2020}, in which CBFs are not trainable, thus, potentially limiting the system's learning performance.  In \cite{Deshmukh2019, Zhao2021, Ferlez2020}, safety guaranteed neural network controllers have been learned through verification-in-the-loop training.  The verification approaches cannot ensure coverage of the entire state space. More recently, CBFs have been incorporated into neural ODEs to equip them with specification guarantees \cite{xiao2022forward}. However, none of these methods can be applied in diffusion models, which we address in this paper.

\section{Conclusions, Discussions and Future Work}

We have proposed finite-time diffusion invariance for diffusion models to ensure safe planning for safety-critical applications. We have demonstrated the effectiveness of our method on a series of robotic planning tasks.
Nonetheless, our method face a few shortcomings motivating for future work.

Specifically, specifications for diffusion models should be expressed as continuously differentiable constraints that may be unknown for planning tasks. Further work may explore how to learn specifications from history trajectory data. There is also a gap between planning and control using diffusion models. We may further investigate diffusion for safe control policies when robot dynamics are known or to be learned. 

\textbf{Broader Impact.} Our proposed finite-time diffusion invariance can be applied to guarantees in other tasks, such as image generations, in addition to planning. In such cases, we can ensure the generation of desired patterns/objects, such as a cat in the generated images. We will further investigate those directions in our future work.

\section{Acknowledgement}
The research was supported in part by Capgemini Engineering. It was also partially sponsored by the United States Air Force Research Laboratory and the United States Air Force Artificial Intelligence Accelerator and was accomplished under Cooperative Agreement Number FA8750-19-2-1000. The views and conclusions contained in this document are those of the authors and should not be interpreted as representing the official policies, either expressed or implied, of the United States Air Force or the U.S. Government. The U.S. Government is authorized to reproduce and distribute reprints for Government purposes notwithstanding any copyright notation herein. This research was also supported in part by the AI2050 program at Schmidt Futures (Grant G-
965 22-63172).

\bibliographystyle{abbrvnat}
\bibliography{MCBF}

\clearpage
\beginsupplement

\section{Proofs}

\subsection{Proof of Theorem 3.2}

\textbf{Proof:} Given a continuously differentiable constraint $h(\bm x_t)\geq 0$ ($h(\bm x_0)\geq 0$), by Nagumo's theorem \cite{Nagumo1942berDL}, the necessary and sufficient condition for the satisfaction of $h(\bm x_t)\geq 0, \forall t\geq 0$ is 
 $$\dot h(\bm x_t) \geq 0, \textit{ when } h(\bm x_t) = 0,$$

If $b(\bm x_k^N)\geq 0, k\in\{0, \dots, H\}$, then 
the condition (\ref{eqn:diff_cbf}) is equivalent to
$$
\frac{db(\bm x_k^j)}{d\bm x_k^j } \dot{\bm x}_k^j + \alpha(b(\bm x_k^j)) \geq 0,
$$
where $\dot{\bm x}_k^j$ is the diffusion time derivative. The last equation is equivalent to
$$\dot b(\bm x_k^j) + \alpha(b(\bm x_k^j)) \geq 0,$$
Since $\alpha$ is an extended class $\mathcal{K}$ function, we have that
$$\alpha(b(\bm x_k^j))\rightarrow 0, \textit{ as } b(\bm x_k^j)\rightarrow 0,$$
In other words, we have $\dot b(\bm x_k^j) \geq 0$ when $ b(\bm x_k^j) = 0$. Since $b(\bm x_k^N)\geq 0, k\in\{0, \dots, H\}$, then by Nagumo's theorem, we have $b(\bm x_k^j)\geq 0, \forall j\in\{0, \dots, N-1\}$. Therefore, the diffusion procedure $p_{\theta}(\bm\tau^{i-1}|\bm\tau^{i}), i\in\{1,\dots, N\}$ is finite-time diffusion invariant, and the finite time in diffusion invariance is $N$. 

If, on the other hand, $b(\bm x_k^N)< 0, k\in\{0, \dots, H\}$, then we can define a Lyapunov function:
\begin{equation}
    V(\bm x_k^j) = -b(\bm x_k^j), k\in\{0,\dots, H\}, j\in\{0, \dots, N\},
\end{equation}
and $V(\bm x_k^N) > 0$.

Since $\alpha$ is an extended class $\mathcal{K}$ function, replacing $b(\bm x_k^j)$ by  $V(\bm x_k^j)$, the condition (\ref{eqn:diff_cbf}) is equivalent to 
$$
\frac{dV(\bm x_k^j)}{d\bm x_k^j } \dot{\bm x}_k^j + \alpha(V(\bm x_k^j)) \leq 0,
$$
which is equivalent to 
$$
\dot V(\bm x_k^j) + \alpha(V(\bm x_k^j)) \leq 0,
$$
Since $\dot V(\bm x_k^j) \leq -\alpha(V(\bm x_k^j)) < 0$, we have that $V(\bm x_k^j)$ will be stabilized to 0 by Lyapunov stability theory. In other words, the state $\bm x_k^j$ will be stabilized to the boundary $b(\bm x_k^j) = 0$. Specifically, when $\alpha$ is a linear function, the last equation can be rewritten as
\begin{equation} \label{eqn:clf1}
\dot V(\bm x_k^j) + \epsilon V(\bm x_k^j)) \leq 0,
\end{equation}
where $\epsilon > 0$.
Suppose we have 
$$\dot V(\bm x_k^j) + \epsilon V(\bm x_k^j) = 0, $$
the solution to the above equation is
$$V(\bm x_k^j) =  V(\bm x_k^N)e^{-\epsilon (N-j)},$$
Using the comparison lemma \cite{Khalil2002}, equation (\ref{eqn:clf1}) implies that
$$
V(\bm x_k^j) \leq  V(\bm x_k^N)e^{-\epsilon (N-j)}, j\in\{0,\dots, N\},
$$
Therefore,
$$
V(\bm x_k^j) \rightarrow 0, \textit{ as } j\rightarrow 0, \textit{ if $N$ is sufficiently large},
$$
and the state $\bm x_k^j$ will be exponentially stabilized to the boundary $b(\bm x_k^j) = 0$. If at diffusion time $j\in\{0,\dots. N-1\}$, the state $\bm x_k^j$ is close to the boundary, and the probability for the state $\bm x_k^j$ to jump into the set: $\{\bm x_k^j: b(\bm x_k^j)\geq 0\}$ is $p$ (due to the Gaussian transitions in the diffusion process), then the probability for  $\bm x_k^0$ to enter the set $\{\bm x_k^j: b(\bm x_k^j)\geq 0\}$ is $1 - (1-p)^j$. When $j$ is large enough, then the probability such that $b(\bm x_k^0)\geq 0$ is almost 1. When $b(\bm x_k^l)\geq 0$ at step $l\leq j$, then $b(\bm x_k^r)\geq 0$ for all $r\leq l$ following the Nagumo's theorem (as in the first case of the proof). Therefore, the diffusion procedure $p_{\theta}(\bm\tau^{i-1}|\bm\tau^{i}), i\in\{1,\dots, N\}$ is finite-time diffusion invariant with almost probability 1. $\blacksquare$

\subsection{Proof of Theorem 3.3}

\textbf{Proof:} Suppose the weight $w_k(j)$ is chosen such that $w_k(j) = 0$ when $j = 0$, then the condition (\ref{eqn:diff_rcbf}) becomes a hard constraint when $j < 0$. In other words, equation (\ref{eqn:diff_rcbf}) becomes:
$$
h(\bm u_k^j|\bm x_k^j):= \frac{db(\bm x_k^j)}{d\bm x_k^j } \bm u_k^j + \alpha(b(\bm x_k^j)) \geq 0, k\in\{0,\dots, H\}, j\in\{-N_a,\dots, 0\},
$$
Then, the proof is simiar to that of the Thm. 3.2, and we have that the diffusion procedure $p_{\theta}(\bm\tau^{i-1}|\bm\tau^{i}), i\in\{-N_a,\dots, N\}$ is finite-time diffusion invariant with almost probability 1. $\hfill \blacksquare$

\subsection{Proof of Theorem 3.4}

\textbf{Proof:} Since $\gamma_k(N) \leq b(\bm x_k^N)$, we have that $s(\bm x_k^j, \gamma_k(j)):= b(\bm x_k^j) - \gamma_k(j) \geq 0$ when $j = N$.

The condition (\ref{eqn:diff_cbft}) is equivalent to 
$$
\frac{\partial s(\bm x_k^j, \gamma_k(j))}{\partial\bm x_k^j } \bm u_k^j + \frac{\partial s(\bm x_k^j, \gamma_k(j))}{\partial j } + \alpha(s(\bm x_k^j, \gamma_k(j))) \geq 0,
$$
which can be rewritten as
$$
\dot s(\bm x_k^j, \gamma_k(j)) + \alpha(s(\bm x_k^j, \gamma_k(j))) \geq 0,
$$

Using the Nagumo' theorem presented in the proof of Thm. 3.2, we have that
$$s(\bm x_k^j, \gamma_k(j)) \geq 0, \forall j \in\{0,\dots, N\}$$
since $s(\bm x_k^N, \gamma_k(N))\geq 0$.

As $\gamma_k(0) = 0$ and $s(\bm x_k^j, \gamma_k(j)):= b(\bm x_k^j) - \gamma_k(j)$, we have that $b(\bm x_k^0)\geq 0, \forall k\in\{0,\dots, H\}$. Therefore,  the diffusion procedure $p_{\theta}(\bm\tau^{i-1}|\bm\tau^{i}), i\in\{0,\dots, N\}$ is finite-time diffusion invariant, and the finite time in diffusion invariance is $0$. $\hfill \blacksquare$

\newpage

\section{Experiment Details}

\subsection{Safe Planning in Maze}
In this experiment, we aim to impose trajectory constraints on the planning path of a maze. The training data  is publicly available from \cite{janner2022planning}, in which initial positions and destinations in maze are randomly generated. The diffusion model is conditioned on the initial positions and destinations.

\textbf{Specifications.} The simple safety specification for the planning trajectory is defined as an super-ellipse-shape obstacle:
\begin{equation}
    \left(\frac{x - x_0}{a}\right)^2 + \left(\frac{y - y_0}{b}\right)^2 \geq 1, 
\end{equation}
where $(x, y)\in\mathbb{R}^2$ is the state on the planning trajectory, $(x_0, y_0)\in\mathbb{R}^2$ is the location of the obstacle. $a > 0, b > 0$. Since the state $(x, y)$ is normalized in diffusion models, we also need to normalize the above constraint accordingly. In other words, we normalize $x_0, a$ and $y_0, b$ according to the normalization of $(x, y)$  along the $x$-axis and $y-$axis, respectively.

The complex safety specification for the planning trajectory is defined as an ellipse-shape obstacle:
\begin{equation}
    \left(\frac{x - x_0}{a}\right)^4 + \left(\frac{y - y_0}{b}\right)^4 \geq 1, 
\end{equation}
We also normalize the above constraint as in the simple case. In this case, it is non-trivial to truncate the planning trajectory to satisfy the constraint. When we have much more complex specifications, it is too hard for the truncation method to work.

\textbf{Model setup, training and testing.} The diffusion model structure is the same as the open source one (Maze2D-large-v1) provided in \cite{janner2022planning}. We set the planning horizon as 384, the diffusion steps as 256 with an additional $N_a = 50$ diffusion steps for the proposed methods. The learning rate is $2e^{-4}$ with $2e^6$ training steps. The training of the model takes about 10 hours on a Nvidia RTX-3090 GPU. More parameters are provided in the attached code: ``safediffuser/config/maze2d.py''. The switch of different (proposed) methods in testing can be modified in ``safediffuser/diffuser/models/diffusion.py'' through ``GaussianDiffusion.p\_sample()'' function.

\begin{figure}[t]
    \centering
    \subfigure{\includegraphics[width=0.24\linewidth]{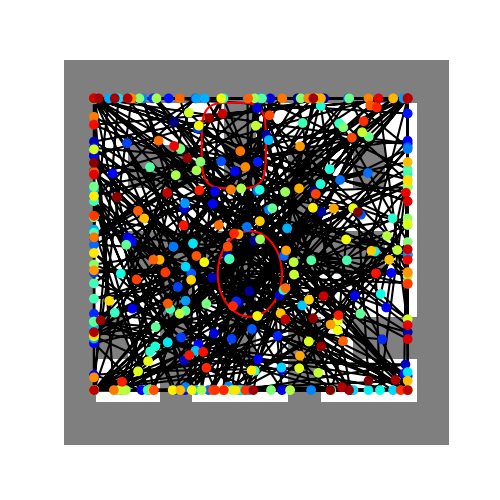}}
    \subfigure{\includegraphics[width=0.24\linewidth]{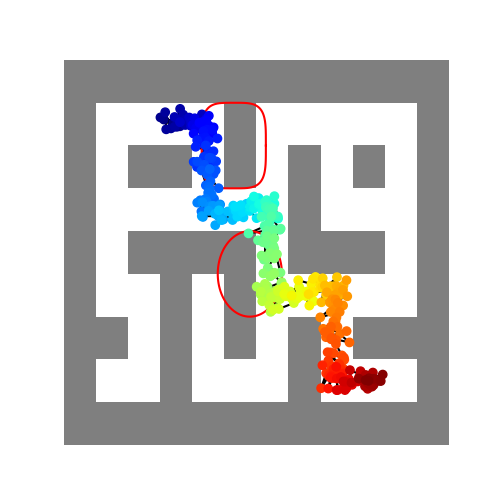}}
    \subfigure{\includegraphics[width=0.24\linewidth]{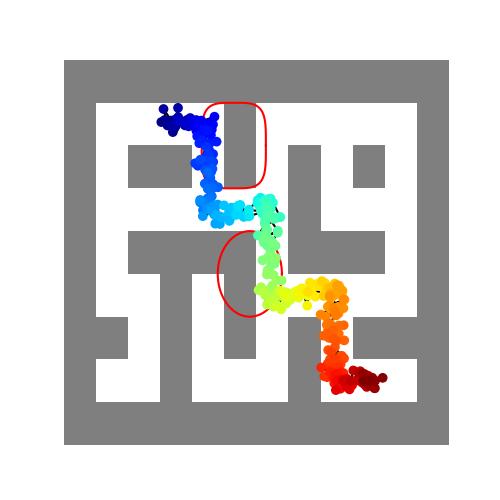}}
    \subfigure{\includegraphics[width=0.24\linewidth]{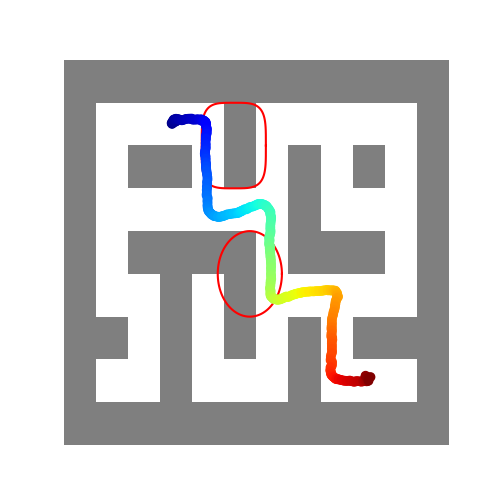}}
    \vspace{-3mm}
    \caption{Maze planning (blue to red) denoising diffusion procedure with diffuser (Left to right: diffusion time steps 256, 4, 3, 0, respectively). Red ellipse and superellise (outside) denote safe specifications. Both specifications are violated with the trajectory from diffuser.
    }
    \label{fig:maze_diffuser}
    \vspace{-5mm}
\end{figure}

\begin{figure}[t]
    \centering
     \subfigure{\includegraphics[width=0.48\linewidth]{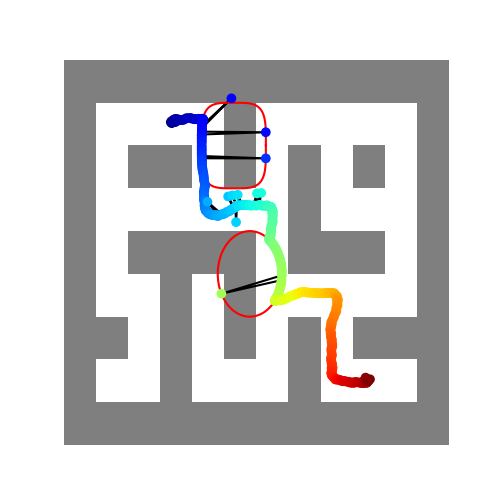}}
    \vspace{-3mm}
    \caption{Maze planning (blue to red) denoising diffusion procedure with robust-safe diffuser at diffusion time step 0. Red ellipse and superellise (outside) denote safe specifications. Although with safety guarantees, some trajectory points may get stuck in local traps.
    }
    \label{fig:maze_diffuser_s}
    \vspace{-5mm}
\end{figure}

In Fig. \ref{fig:maze_diffuser}, we present a diffusion procedure using the diffuser, in which case the generated trajectory can easily violate safety constraints. Using the proposed robust-safe diffuser, the generated trajectory can guarantee safety, but some points on the trajectory may get stuck in local traps, as shown in \ref{fig:maze_diffuser_s}. Using the proposed relaxed-safe diffuser and time-varying-safe diffuser, the local trap problem could be addressed.

\subsection{Safe Planning for Robot Locomotion}

For robot locomotion (in MuJoCo), we wish the robot to avoid collisions with obstacles, such as the roof. In this case, since there is no local trap problem, we only consider robust-safe diffuser (RoS-diffuser). Others work similarly. The training data set is publicly available from \cite{janner2022planning}.

\textbf{Specifications.} The simple safety specification for both the Walker2D and Hopper is collision avoidance with the roof. In other words, the height of the robot head $z\in\mathbb{R}$ should satisfy the following constraint:
\begin{equation}
    z \leq h_r,
\end{equation}
where $h_r > 0$ is the height of the roof. We also need to normalize $h_r$ according to the normalization of the state $z$ in the diffusion model.

The complex safety specification for both the Walker2D and Hopper is a speed-dependent collision avoidance constraint:
\begin{equation}
    z + \varphi v_z \leq h_r,
\end{equation}
where $\varphi > 0$, $v_z \in\mathbb{R}$ is the speed of the robot head along the $z$-axis. The speed-dependent safety constraint is more robust for the robot to avoid collision with the roof since when the robot jumps faster, we need to ensure a larger safe distance with respect to the roof in order to account for all kinds of uncertainties or perturbations. In this case, the simple truncation method is hard to work since it is not clear how to truncate both $z$ and $v_z$ at the same time.

\textbf{Model setup, training and testing.} The diffusion model structures are the same as the open source ones (Walker2D-Medium-Expert-v2 and Hopper-Medium-Expert-v2) provided in \cite{janner2022planning}. We set the planning horizon as 600, the diffusion steps as 20. The learning rate is $2e^{-4}$ with $2e^6$ training steps. The training of the model takes about 16 hours on a Nvidia RTX-3090 GPU.  More parameters are provided in the attached code: ``safediffuser/config/locomotion.py''. The switch of different methods in testing can be modified in ``safediffuser/diffuser/models/diffusion.py'' through ``GaussianDiffusion.p\_sample()'' function.

\subsection{Safe Planning for Manipulation}

For manipulation (in Pybullet), the diffusion models generate joint trajectories (as controls) for the robot, which are conditioned on the locations of the objects to grasp and place. The training data set is publicly available from \cite{janner2022planning}. Specifications are joint limitations to avoid collision in joint space.

\textbf{Specifications.} The simple safety specification for the robot is in the joint space, and we are trying to limit the joint angles of the robot within allowed ranges:
\begin{equation}
    x_{min}\leq x\leq x_{max}, 
\end{equation}
where $x \in\mathbb{R}^7$ is the state of 7 joint angles, $x_{min} \in\mathbb{R}^7$ and $x_{max} \in\mathbb{R}^7$ denotes the minimum and maximum joint limits.
We need to normalize the limits according to how the state $x$ is normalized in the diffusion model.

The complex safety specifications are speed-dependent joint constraints:
\begin{equation}
    x_{min}\leq x + \varphi v\leq x_{max}, 
\end{equation}
where $\varphi>0$, $v\in\mathbb{R}^7$ is the joint speed corresponding to the joint angle $x$. In this example, since the diffusion model does not directly predict $v$, we evaluate $v$ using $x(k)$ and $x(k+1)$ along the planning horizon. The joints limits are also normalized as in the simple specification case.

\textbf{Model setup, training and testing.} The diffusion model structure is the same as the open source one provided in \cite{janner2022planning}, and we use their pre-trained models to evaluate our methods when comparing with other approaches.

\end{document}